\documentclass[10pt, conference]{IEEEtran}

\usepackage{amsmath,amssymb,amsfonts}
\usepackage{graphicx}
\usepackage{pifont}

\usepackage{amsthm}
\usepackage{xcolor}
\usepackage[figuresright]{rotating}

\usepackage{graphicx}
\graphicspath{{/}{fig/}}

\usepackage{array}
\usepackage{textcomp}
\usepackage{xcolor}
\usepackage{multirow}
\usepackage{booktabs}

\usepackage{mathtools}
\usepackage{breqn}
\usepackage{float}

\usepackage{pgfplots}
\pgfplotsset{compat=1.7}
\usepgfplotslibrary{groupplots}

\usepackage{caption}
\usepackage{subcaption}

\usepackage{multirow,tabularx}
\usepackage{hyperref}
\usepackage{flushend}
\usepackage{algorithmic}
\usepackage[vlined, ruled, shortend]{algorithm2e}

\newlength\figureheight
\newlength\figurewidth
\SetAlCapNameFnt{\footnotesize}
\SetAlCapFnt{\footnotesize}

\title{
    Distributed Robotic Systems in the Edge-Cloud Continuum with ROS\,2: a Review on Novel Architectures and Technology Readiness
}

\author{
    \IEEEauthorblockN{
        \vspace{1em}
        Jiaqiang Zhang\IEEEauthorrefmark{2},
        Farhad Keramat\IEEEauthorrefmark{2},
        Xianjia Yu\IEEEauthorrefmark{2}, 
        Daniel Montero Hern\'andez\IEEEauthorrefmark{2},\\[-.8em]
        Jorge Pe\~na Queralta\IEEEauthorrefmark{2},
        Tomi Westerlund\IEEEauthorrefmark{2} \\[+1em]
    }
    \IEEEauthorblockA{
        \normalsize
        \IEEEauthorrefmark{2}\href{https://tiers.utu.fi}{Turku Intelligent Embedded and Robotic Systems (TIERS) Lab, University of Turku, Finland}.\\
        Emails: \textsuperscript{1}\{jizhan, fakera, xianjia.yu, daanmh, jopequ, tovewe\}@utu.fi\\[+6pt]
    }
}

\begin{document}

\maketitle
\thispagestyle{empty}
\pagestyle{empty}



\begin{abstract}%
    \label{sec:abstract}%
    Robotic systems are more connected, networked, and distributed than ever. New architectures that comply with the \textit{de facto} robotics middleware standard, ROS\,2, have recently emerged to fill the gap in terms of hybrid systems deployed from edge to cloud. This paper reviews new architectures and technologies that enable containerized robotic applications to seamlessly run at the edge or in the cloud. We also overview systems that include solutions from extension to ROS\,2 tooling to the integration of Kubernetes and ROS\,2. Another important trend is robot learning, and how new simulators and cloud simulations are enabling, e.g., large-scale reinforcement learning or distributed federated learning solutions. This has also enabled deeper integration of continuous interaction and continuous deployment (CI/CD) pipelines for robotic systems development, going beyond standard software unit tests with simulated tests to build and validate code automatically. We discuss the current technology readiness and list the potential new application scenarios that are becoming available. Finally, we discuss the current challenges in distributed robotic systems and list open research questions in the field.
\end{abstract}

\IEEEpeerreviewmaketitle



\section{Introduction}\label{sec:introduction}

With robotic systems being increasingly connected, there is a growing intersection between the IoT and robotics domains~\cite{simoens2018internet}. Specifically, much potential can be found at the intersection of autonomous robots with the edge and cloud domains~\cite{queralta2020reconfigurable, saha2018comprehensive}. This is leading to a myriad of new architectures, technologies and proposals for robots to leverage the edge-cloud computing continuum to enhance autonomy~\cite{queralta2020enhancing}, drive multi-robot cooperation and human-robot collaboration~\cite{huang2022edge, fosch2019cloud, chen2018study}, and provide additional computational capabilities~\cite{huaimin2018cloud}.

\begin{figure}
    \centering
    \includegraphics[width=.48\textwidth]{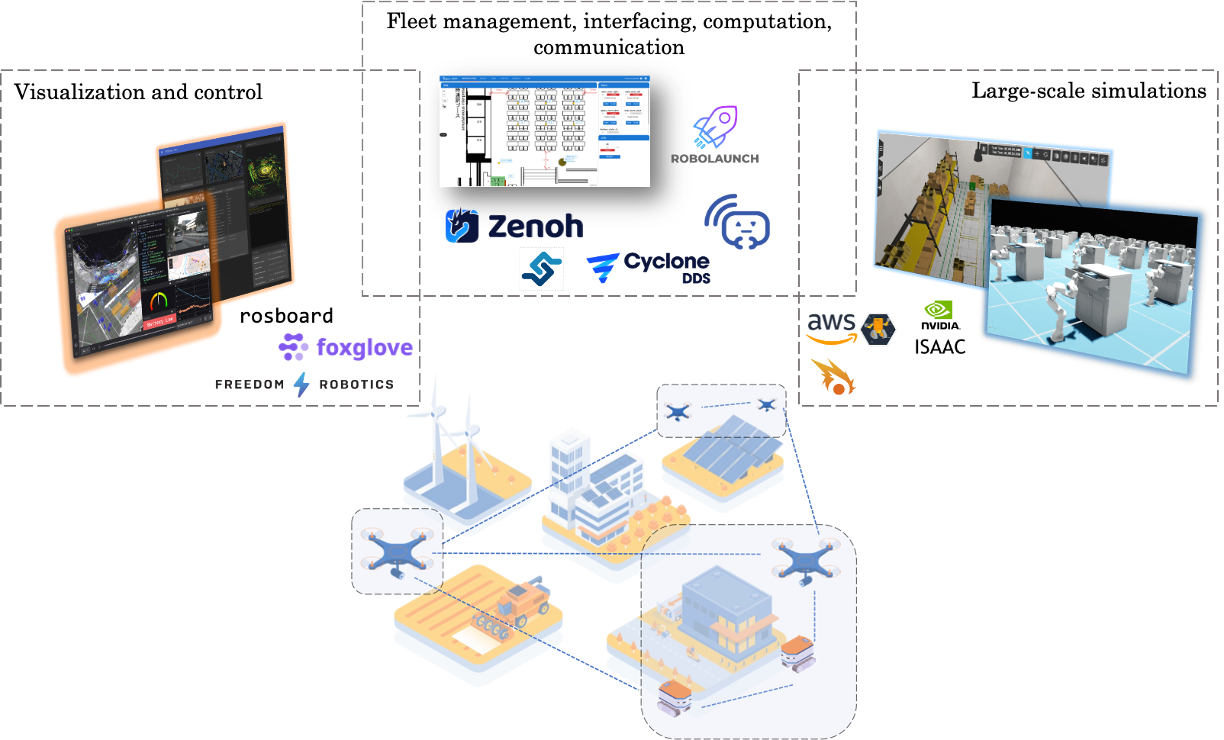}
    \caption{Illustration of a subset of edge-fog-cloud robotic platforms and tools, and the potential areas where cloud and edge computing can play a key role for autonomous robots. From visualization and teleoperation to simulation and online learning, and including fleet management or computational offloading, connectivity can be leveraged to build more robust and capable distributed robotic systems.}
    \label{fig:concept}
\end{figure}

The benefits of wider usage of edge and cloud computing paradigms and solutions to robotic systems are indeed enormous; however, widespread adoption requires further integration into the existing tools and frameworks, and particularly with the Robotic Operating System (ROS), the \textit{de facto} standard middleware for the design and development of robotic systems. Therefore, even though there is an extensive literature in the areas of edge, fog, and cloud robotic systems, only recently more general frameworks.

From the perspective of users, we are witnessing a new wave of remote visualization and teleoperation tools that are cloud-native, mobile-ready, and ROS-based~\cite{dawarka2022building, openrmf}. At the same time, a rapidly increasing number of simulation tools for both desktop and cloud are enabling next-generation embodied AI and robot learning algorithms~\cite{shah2018airsim, makoviychuk2021isaac}. These simulators are leveraging physics and game engines leading to more photorealistic and larger-scale simulations~\cite{szot2021habitat, yuan2021presim, nvidia-sim}. In the cloud, simulations are also being connected to continuous integration and development pipelines (CI/CD)~\cite{toffetti2020cloud}, and to the Metaverse and digital twin platforms~\cite{li2022exploring}. In parallel to all of these, new ROS extensions allow for easily switching between edge and cloud ROS deployments~\cite{chen2021fogros, ichnowski2022fogros}, and a myriad of platforms for fleet management and multi-robot cooperation are emerging~\cite{openrmf}.

Within the above context, this review paper covers novel design approaches, architectures, and technologies that leverage the increasing connectivity and ubiquity of robots. In contrast to other surveys and reviews in the literature, we look closely at the compliance of such systems and architectures with ROS. We delve into the basic concepts behind these technologies and frameworks (containerization, computational offloading, and security, among others), and into the specific technologies driving the state-of-the-art ROS integrations in the edge-cloud continuum. We then describe different use cases in the literature that demonstrate the potential of these technologies across industries and application areas.


The remainder of this document is organized as follows. Section II introduces background concepts related to ROS\,2, DDS, containerization, DLTs, and computational offloading orchestration. Section III then reviews novel architectures and technologies for autonomous robots in the edge-cloud continuum, with Section IV focusing on application scenarios and use cases for the different technologies. Section V discusses open research questions and Section VI concludes the work.




\section{Background}
\label{sec:related_work}

Through this section, we introduce the key technologies and concepts behind today's autonomous and distributed robotic systems, but also those with the potential to play a more important role in the next generation of technologies (e.g., DLTs).

\subsection{ROS\,2 and DDS}


The Robot Operating System (ROS) is the \textit{de-facto} standard in autonomous robotic systems. ROS\,2 is a revamped version of the original ROS middleware that has been under development for over half a decade. ROS\,2 design principles include (i) distribution, (ii) abstraction, (iii) asynchrony, and (iv) modularity~\cite{macenski2022robot}. All of these, but specially distribution, modularity, and asynchrony, are tightly related to the new reality of connected and decentralized robotic systems, moving from single robots to fleets, and moving from single-host computing to computing architectures naturally distributed across the edge-cloud continuum. At the same time, the increasing connectivity and distribution bring new challenges, with ROS\,2 design requirements including (i) built-in security, and the ability to support (ii) embedded systems, (iii) diverse networking conditions, and (iv) real-time computing. A final design requirement is (v) product readiness. Again, these extend the natural requirements of robust system architectures in the edge-cloud continuum, with a need for built-in security and resilience against alternating network conditions, as well as support for heterogeneous computing hardware. 


A typical ROS\,2 system is composed of a number of distributed processes or \textit{nodes}, ranging from sensor drivers to algorithm implementations, high-level decision-making, or external interfacing and control. ROS\,2 nodes communicate to each other through available communication patterns: (i) topics, an asynchronous message passing framework with a pub/sub system; (ii) services, synchronous request-response patterns; and (iii) actions, asynchronous interfaces with a request-feedback-response architecture that fits particularly well physical actions requiring time and physical interaction with a robot's environment. Additionally, ROS\,2 provides a series of abstraction layers, with the communication middleware relying on implementations for the industry-standard DDS protocol~\cite{profanter2019opc}. In the different architectures reviewed in this paper, we will look at how some use-cases require the use of additional communication middlewares or data streaming possibilities. This is owing to the data-intensive DDS middleware and the high-density and uncompressed raw data provided by some sensors (e.g., high-quality images or three-dimensional point cloud) not allowing for offloading computation from edge to cloud directly.

\subsection{Containerization}


Virtualization technologies are some of the core foundations of cloud computing. Virtual machines (VMs) have been used as the backbone of cloud computing to provide virtualized operating systems, until the outbreak of container technology~\cite{pahl2015containerization}.
Containers offer a lightweight virtualization solution where the application can run on the environment with all the dependencies required.
Docker and Kubernetes have revolutionized virtual techniques and cloud computing, and both are leading container orchestration tools nowadays~\cite{uphill2017devops, armstrong2016devops}.
Docker is a commercial containerization platform and runtime where researchers and developers can build, deploy, and run containers.
Kubernetes, which was originally developed by Google, is a platform to orchestrate containers~\cite{shah2019building}.

\subsection{Computational Offloading}


In its most basic form, it can be argued that a cloud robotics framework must allow for either (i) remote teleoperation of robots, (ii) running part of the computation required for autonomous operation in the cloud, or (iii) both of those features. This leads to computational offloading being one of the main studied areas in the literature of cloud robotics~\cite{afrin2021resource}.
At the same time, the standardization of the abstract computing distribution paradigm that edge computing defines brought forward by the European Telecommunications Standards Institute (ETSI) with multi-access edge computing (MEC) is built around computational offloading possibilities to the radio access network (RAN)~\cite{etsi2018mec}. Indeed, with the advent of 5G, 6G and, beyond mobile networking solutions, low-latency and high-throughput computational offloading is made possible by using computational resources of servers deployed across the RAN layer. The "multi-access" term puts an emphasis on the multi-tenant infrastructure, meaning that such computing infrastructure is more limited than in the cloud and therefore it is more important to manage, or \textit{orchestrate}, how it is shared between services. In next-generation robotic systems, computational offloading has the potential to become a more broad and complex area as computation is distributed across the edge, between edge and cloud, but also anywhere in between (e.g., the RAN layer) within a heterogeneous computing continuum. Key parameters taken into account in computational offloading orchestration include real-time latency and bandwidth~\cite{queralta2020blockchainpowered}.

\subsection{Security}


SROS2 provides a set of tools, and libraries for providing security, authentication, and access control for ROS\,2 entities~\cite{mayoral2022sros2}. As ROS\,2 is built on top of DDS, the DDS security specification is utilized to implement the security tools for ROS 2 at the communication level. OMG's DDS security specification~\cite{dds2017sec} enforces security models in five Service Plugin Interfaces (SPI). SROS2 implements only three of these SPIs, which are authentication, access control, and cryptography. The authentication SPI enables ROS\,2 entities to establish identities through the use of Public Key Infrastructure (PKI). The access control SPI restricts access to ROS\,2 publish/subscribe communication channels only to predefined authenticated identities. The cryptography SPI encompasses cryptography functions such as encryption, decryption, digital signatures, and hash for the other SPIs.

\subsection{Distributed Ledger Technologies}


Consensus has always been an essential problem in distributed systems~\cite{fischer1983consensus}. Therefore, decentralized consensus plays a key role in building towards robust solutions for the edge-cloud continuum. While earlier research on consensus was restricted to crash tolerance or a limited number of participants~\cite{charron2001agreement}, with the advent of distributed ledger techniques (DLT), research on this domain has advanced~\cite{bano2019sok}. DLTs provide an immutable ledger of the messages transmitted between the participant nodes. The shared immutable ledger facilitates consensus and provides a secure medium for the interaction of the participants in a distributed system. By using a DLT, trust is removed from third parties, resulting in more decentralized systems. Apart from decentralization, DLTs also offer security features, making them important components of distributed system development~\cite{gorbunova2022distributed}. Distributed robotics systems are among the areas where DLTs are being used due to their need for decentralized agreement solutions and secure communication channels~\cite{castello2018blockchain}. Blockchain solutions have already been applied to different robotic use-cases in the literature, but newer technologies such as those using directed acyclic graph (DAG) architectures are paving the way for more scalable and lightweight DLTs that have significant potential in distributed robotic systems~\cite{salimi2022towards, salimi2022secure, keramat2022partition, salimpour2022decentralized, torrico2022uwb}.

Among the key technologies used in the literature are Ethereum~\cite{dannen2017introducing}, Hyperledger Fabric~\cite{androulaki2018hyperledger}, and IOTA~\cite{bartolomeu2018iota}. An overview of consensus algorithms and use cases within the more general edge-cloud domain can be found in~\cite{queralta2021blockchainmec}. A subset of those consensus algorithms is reviewed within the context of robot swarm systems in~\cite{strobel2020blockchain}.



 

\section{Technology readiness}
\label{sec:technology}


This section covers the latest advances in cloud robotics with ROS\,2, DLT integrations to ROS\, 2, containerization and edge-to-cloud system architectures. In Table~\ref{tab:technology}, we list some of the state-of-the-art robotics platforms or tools applied on the cloud and edge.

\begin{table*}[]
    \centering
    \caption{Selection of platforms, technologies, and tools in the robotics domain for the edge-cloud continuum.}
    \label{tab:technology}
    \footnotesize
    \begin{tabular}{@{}lcccl@{}}
        \toprule
        Technology & Computing & Role/Type & Open-Source & Description \\ \midrule
        \textbf{FogRos2}\,\cite{ichnowski2022fogros} & Edge-Cloud & Deployment & Yes & Integrated to ROS\,2 launch system\\
        \textbf{Foxglove} & Edge+Cloud & Data recording, visualization & Partly & MCAP format alternative to standard rosbag in ROS\,2  \\
        \textbf{Robolaunch} & Cloud & Simulation, deployment & Partly & End-to-end infrastructure for development and simulation \\
        \textbf{NVIDIA Omniverse} & Edge-Cloud & Ecosystem & Partly & Simulation, fleet management, HW acceleration, and others. \\
        \textbf{ROBOTCORE} & Edge & Hardware acceleration & Yes & Reference hardware architecture for edge acceleration. \\
        \textbf{ROSBoard} & Edge & Visualization & yes & Lightweight web server for visualization and teleoperation. \\
        \textbf{AWS RoboMaker} & Cloud & Simulation, deployment & Partly & Automated tests, scriptable simulations, end-to-end integration. \\
        \textbf{Open-RMF} & Edge-Cloud & Fleet management & Yes & Multi-tenant, interoperable, fleet management operations. \\ 
        \textbf{Formant} & Cloud & Fleet management & Partly & Platform for visualization and fleet management \\
        \textbf{Vulcanexus} & Edge-Cloud & ROS\,2+DDS Platform & Partly & ROS\,2 tool set platform with cloud integrations. \\
        \textbf{Zenoh} & Edge + Cloud &  Communication middleware & Yes & Discovery-efficient pub-sub communications. \\
        \bottomrule
    \end{tabular}
\end{table*}

\subsection{ROS\,2 in the Cloud}



Cloud computing has been playing an increasingly important role in the last few years in robotics, owing in part to the increase in the number of embedded devices and the availaibility of high-speed networking solutions~\cite{mehrooz2019system, lumpp2021container}. 

As an example, Amazon AWS Robomaker~\cite{robomaker} is a cloud-based platform for developing, testing, deploying intelligent robotics applications, and running simulations. Robotics developers can scale and automate simulation workloads, run large-scale and parallel simulations with Amazon AWS's technologies on cloud computing, big data processing, and web services.
Furthermore, AWS RoboMaker can run automated test within a CI/CD pipeline, train reinforcement models, and test multi-robot system.


An alternative ecosystem is NVIDIA Omniverse. The NVIDIA Isaac Sim~\cite{nvidia-sim} powered by Omniverse, provides photorealistic, physically accurate environments for developing, testing, and management of AI-based robots with sim-to-real capabilities. Regarding training perception models, Isaac Sim can generate synthetic data with Isaac Replicator allowing for adjustments in texture, colors, illumination, and location. It is available both locally and on the cloud for scalability. Notably, NVIDIA Isaac Sim is compatible with ROS and ROS2 and supports a variety of sensor modalities. ROS developers can move simulations from the most popular simulator in robotics, Gazebo, to Isaac Sim by exploiting the connectors between Gazebo and Isaac Sim. In addition, Isaac Sim is capable of real-time collaboration with teams from around the world for the development of AI robots. NVIDIA Isaac Sim can execute vision-based deep learning tasks including object detection, segmentation, spatial detection, Reinforcement Learning tasks provided by NVIDIA Isaac Gym, and other similar tasks.

\subsection{Edge-to-Cloud ROS\,2 Integrations}

While cloud computing has become increasingly popular, computing has been extended from the cloud side to the edge side to decrease latency and network congestion. In addition to the cloud and edge paradigms, fog computing provides computing resources with high quality and low latency at the edge-side network, aiming to bring processing capabilities closer to the consumers, prevent overuse of cloud resources, and reduce operating costs~\cite{sabireen2021review}. 

One of the most prominent platforms supporting the edge-to-cloud deployment of ROS\,2 nodes is FogROS2~\cite{ichnowski2022fogros}. FogROS2 extends the previous version of the platform, FogROS~\cite{chen2021fogros}, tightly integrating with the ROS\,2 launch and communication system. It allows for existing ROS\,2 nodes to be deployed either in the cloud or at the edge, and then adapts the communication patterns accordingly to ensure real-time data delivery. The FogROS2 framework relies on a VPN connection to secure the ROS\,2 DDS communication while allowing for the auto-discovery mechanisms in DDS to seamlessly bridge edge and cloud nodes. 
To allow for image processing on the cloud side for data generated at the edge, FogROS2 relies on H.264, a compression method used to reduce the bandwidth and latency and compress successive images in video format rather than individually (the default ROS compression).  The H.264 encoding is done before publishing to the corresponding topic; the H.264 decoding is done before the subscribers receive the information. Also, a browser-based 3D visualizer, Foxglove, is used for remote monitoring and visualization. According to the paper, FogROS2 supports AWS and GCP. FogROS2 is evaluated on 3 example robot applications where cloud computing accelerates processing greatly compared to the computing capabilities of the robot and shows a significant performance benefit to using cloud computing, with the additional improvement from transparent video compression.


Other works in the literature have focused instead on improving the containerization of ROS nodes to enable more widespread adoption. For example, Bakshi et al. present a discussion on key considerations for the integration of ROS applications with container-based architectures~\cite{bakhshi2021fault}. Using Fault Tree Analysis, the authors identify the limitation of existing container-based virtualization architectures, the lack of persistent storage for stateful applications at the fog layer, which can be overcome by applying storage containers at the edge side.


\subsection{ROS\,2 in Containers}


Containers and DDS have been applied to solve the challenges caused by the growing number of edge devices and demand for low latency. Some container-based methodology for drones has been explored. At first, the containers are only used on the cloud side. In the paper~\cite{mehrooz2019system}, the authors propose an open-source drone-cloud platform based on Service-Oriented Architecture (SOA) and ROS. A Kubernetes cluster is deployed as cloud services on the first layer of a three-layer framework. The second and the third layer are ROS/ROS\,2 and the simulator or flight controller.


Containerized services are used to assure the robustness of autonomous systems. In~\cite{matlekovic2022microservices}, the authors use containerized simulation to assure cloud-to-UAV communication. Xfce desktop environment and Virtual Network Computing(VNC) are used in the UAV simulation. The authors build a docker image with Dockerfile and run it on a Kubernetes cluster. In order to deploy the applications automatically, they use a Kubernetes cluster with three nodes to manage the docker images.

With a lightweight container platform such as K3S, edge-side applications can also be containerized. For instance, Lumpp et al.~\cite{lumpp2021container} propose a design methodology that containerizes and orchestrates ROS applications for heterogeneous and hierarchical hardware architectures. 
An extension of the Docker and K3S which is a lightweight Kubernetes distribution was used to build a Cloud-Edge platform that could deploy ROS-based robotic applications.

\subsection{DLT Integrations}

The vast majority of experimental applications and use-cases in the literature involving DLTs and autonomous robotic systems are implemented through the integration of the Ethereum blockchain and ROS\,1. The first few works in the literature integrating blockchain for robots were mostly related to the swarm robotics domain~\cite{castello2018blockchain, strobel2018managing, dorigo2018blockchain}. For example, in~\cite{castello2018blockchain}, multiple potential integrations and use cases are introduced for a variety of robots and blockchain uses (e.g., as a marketplace and cryptocurrency platform, or as a collaborative decision-making platform). A specific application of byzantine fault-tolerant consensus algorithms in blockchain solutions is the detection of byzantine agents within a robot swarm or multi-robot system. An early implementation of such an approach in~\cite{strobel2018managing} shows the potential of Ethereum smart contracts for identifying agents that are fabricating data or altering their sensor data in a cooperative mission. Other approaches in the literature include an architecture for building a shared \textit{reputation} measure based on the quality or compliance of sensor data~\cite{dorigo2018blockchain}, as well as different approaches to collaborative decision-making processes~\cite{singh2020efficient}.


More recently, however, new architectures have emerged that use-next generation permissioned blockchain frameworks and DAG-based architectures such as IOTA's. For example, a framework for integrating Hyperledger Fabric Smart contracts with ROS\,2 is presented in~\cite{salimi2022towards}. The high throughput of permissible blockchain frameworks makes them suitable for robotic systems, as shown by this integration. Furthermore, the paper discusses that Hyperledger Fabric has the potential to be used to provide identity management, immutable data logs, and secure communication channels for distributed robotic systems without drastically decreasing performance, and even near-real-time applications could be supported. Similarly, integration with the recently launched IOTA Smart Contract Platform in~\cite{keramat2022partition} demonstrates the first ROS and blockchain integration that is tolerant to network partitions. The partition tolerance, in addition to byzantine tolerance, ensures the platform is usable in real-world networks where consistent connectivity cannot always be ensured.


Furthermore, a framework that exploits the functionality and speed offered by Hyperledger Fabric Smart Contracts to aid with the selection of Edge Nodes for Computational Offloading. The design and implementation of a system that analyzes the latency and resources in a network of nodes through blockchain smart contracts in~\cite{montero2022dltedge} demonstrate that the technology can serve dependable distributed applications, and with the addition of execution of containerized applications, could perform operations autonomously.

\subsection{Security}


A methodology for utilizing SROS2 on robotic distributed systems is presented by~\cite{mayoral2022sros2}. This methodology is iterative, starting with threat modeling, identity authentication, generating security requirements, and monitoring deployment. The iteration continues until all required communication channels have been secured. \cite{kim2018security} have conducted experiments on evaluating how enabling ROS\,2 security affects the performance metrics such as latency and throughput of the ROS\,2 communication channel. The experiments are also carried out on wired and wireless networks, as well as Virtual Private Networks (VPN). The results of the experiment could be used to determine how the ROS\,2 security parameters should be configured to achieve the desired performance metrics. They have also performed a security analysis on the implementation of SROS2, and they have claimed that the implementation did not conform with the DDS security specification. SROS2 is still under development and the next versions would resolve the implementation compromises.


In addition, the to use of DLTs, multiple works in the literature have been focusing on securing ROS, ROS\,2, and distributed robotic systems in general. In~\cite{ferrer2021secure}, the authors introduce an approach to \textit{secret} robot missions leveraging Merkle trees for data validation, a structure typically used in blockchain frameworks. The paper presents an architecture for robot swarms to achieve collaboration and cooperative missions without revealing to each other the content of the sub-tasks that are individually performed, but still allowing their validation by other robots. These missions are extended to more generic applications in~\cite{queralta2022secure}, where the authors focus on instruction graphs that can be iteratively validated but are unknown when a mission starts.

\subsection{Distributed machine learning frameworks}


With the improvement of ML techniques and the capability to leverage hardware acceleration over the past decade, however, it is still challenging for computation-constrained edge devices to fully exploit the benefits of ML and other artificial intelligence algorithms. The modern era has witnessed the incredible proliferation of robotic edge devices as part of the Internet of Things. This situation demands high-level collaboration among those robots situated in various locations across the globe. 

A distributed machine learning framework was brought forward to utilize the multiple processors to accelerate the DL training process by parallel the computation and the data~\cite{xing2015petuum, tang2020communication}. 
A substantial quantity of literature is devoted to distributed DL with the purpose of achieving tighter collaboration between cloud and edge computing~\cite{wu2020collaborate, jiang2019distributed}. 
This equilibrium between the two paradigms is projected to become more ubiquitous as the IoT age matures. Data security and user privacy are immediate problems that arise with the implementation of distributed DL across the cloud and edge. Through~\cite{shi2021towards, li2020toward, buniatyan2019hyper}, various research directions have emerged with the goal of making distributed learning processes more scalable, secure, and privacy-preserving.
Additionally, other research focuses on employing distributed DL for processing and learning from sensitive data such as health data~\cite{vepakomma2019reducing}
or public institutions~\cite{balachandar2020accounting}.

In the distributed multi-robot system, Federated Learning (FL) has garnered attention~\cite{xianjia2021federated}. Instead of collecting sensor data from numerous edge devices, federated learning maintains the data locally by sharing the trained models at the edge in a decentralized manner in order to create a global model. By doing so, FL ensures the system's security and privacy while maintaining its distributed nature. Due to the heterogeneous computing and communication capabilities of edge devices, fog learning was proposed to dynamically distribute ML model training throughout the continuum of nodes, from edge devices to cloud servers~\cite{hosseinalipour2020federated}. 

\section{Application Scenarios}
\label{sec:use_cases}



This section covers example applications of cloud robotics, and ROS 2 integration to DLTs and edge-cloud systems.

\subsection{Edge-to-Cloud ROS Applications}

Containers have shown great potential in Edge-Cloud ROS\,2 integrations. Docker and  Kubernetes are both widely used container orchestration tools. In~\cite{haiyan2021research}, the authors propose a hybrid cloud-based robot grabbing system, which is deployed on a Kubernetes cluster provided by Alibaba where each ROS node runs in separate docker containers. 
Alternatively, in~\cite{farnham2021umbrella}, the authors propose a collaborative robotics testbed platform based on Docker. 
The testbed platform is deployed on a K3S. %

Some blockchain-based applications can also apply to an Edge-Cloud system. For example, in~\cite{zhou2021blockchain}, the authors propose a decentralized volunteer edge cloud architecture for real-time IoT applications. Kubernetes and KubeEdge are used to integrate smart contracts into the off-chain Edge-Cloud system.

Other works have discussed the potential of ROS\,2 for the next generation of robot swarms~\cite{queralta2021towards}, and in blockchain technology for managing the interaction of robots within heterogeneous swarms of robots~\cite{queralta2020blockchainpowered}.

Looking into the application of one of the most mature frameworks for cloud support, FogROS~\cite{ichnowski2022fogros}, the authors of the paper show the potential for offloading image processing applications, which can then leverage more GPU or even FPGA resources in the cloud. This ties together with the recent efforts in driving standardization for hardware acceleration in ROS~\cite{mayoral2022robotcore, mayoral2021adaptive}.

\subsection{ROS\,2 and Distributed Learning}

Distributed learning augmented by ROS2 can be utilized in a variety of robotic and autonomous system applications. Federated Learning has been demonstrated to be an effective approach for training and deploying vision-based obstacle avoidance models, both in the real world and when simulating photorealistic data~\cite{xianjia2022federated}. Compared to the centralized technique for obstacle avoidance in navigation tasks, it can achieve comparable performance~\cite{xianjia2022towards}. There are multiple research works on enabling Deep Reinforcement Learning in a distributed manner with the help of ROS2 for both the simulation and real-world application ~\cite{lucchi2020robo,lopez2019gym,leal2020fpga}.

\subsection{ROS\,2 and DLT Use Cases}

The integration of DLTs to ROS-based systems has immediate applications in the area of byzantine agent detection~\cite{strobel2018managing}, but specially within the more general of collaborative decision making. In that direction, the literature includes examples for cooperative mapping~\cite{keramat2022partition}, where fabricated or noisy data is eliminated effectively by assessing the compliance and disparity between the maps submitted by different robots. An extension of that work in~\cite{salimpour2022decentralized} shows that smart contracts combined with vision-based change and anomaly detection can also lead to more varied byzantine agent detection methods.



\section{Discussion}
\label{sec:discussion}

Throughout the paper, we have reviewed recent advances in cloud and edge robotics. The number of frameworks, tools, and platforms that have been recently released and are currently under development is staggering. Indeed, robots are no longer independent entities but are instead deployed in fleets and benefit from computing resources across the edge cloud continuum. While much has been improved, there is a number of open research challenges that both the IoT and robotics communities will be potentially solving in the near future.

\subsection{Open Research Questions}

A number of technologies and methods have the potential to play a key role in solving some of the following open research questions:

\textit{Live migration:} how to effectively address the live migration of containerized ROS\,2 nodes between edge and cloud to ensure minimal loss of data? Compared to existing methods, live migration in robotic applications requires addressing real-time requirements and latency constraints, leading to safety concerns if a container migration affects autonomy~\cite{chen2021fogros, ichnowski2022fogros}.

\textit{Interoperability:} the Open-RMF standard is driving multi-fleet interoperability~\cite{openrmf}; however, what other standards and ROS integrations are needed to drive the adoption of Open-RMF?

\textit{Elasticity:} going beyond live migration, how can decentralized workload distribution be adapted to dynamically changing robotic systems, changes in their network topology, and changes in their operational requirements? Elastic computing techniques can bring more robustness and resilience to multi-robot systems, optimizing the computational resources, but multiple challenges remain as these systems have a higher degree of dynamism than more typical cloud or edge clusters~\cite{queralta2020reconfigurable}.

\textit{Decentralization:} in this survey, we have covered decentralized decision-making through DLTs. The research in the area is however still incipient. There is significant potential to combine the decentralized consensus approaches in DLTs with the already distributed nature of ROS\,2. 

\textit{Robustness and enhanced autonomy:} the computing resources available across edge and cloud can enable higher degrees of autonomy and robustness, for example with larger deep learning models in the cloud. However, this also needs additional fallback mechanisms to ensure that autonomy does not depend on stable connectivity to a remote host.

\section{Conclusion}\label{sec:conclusion}

In this paper, we have given a description of novel design approaches, architectures, and technologies that leverage the increasing connectivity and ubiquity of robots. We have reviewed a selection of technologies, tools and frameworks that integrate with ROS\,2 and allow for robots to better exploit the computational resources of the cloud-edge continuum. These technologies are introduced with a series of background concepts, and we look into application scenarios and use cases. We also discuss some of the current challenges and potential directions for future research.


\section*{Acknowledgment}

This research work is supported by the Academy of Finland's RoboMesh project (Grant No. 336061) and AeroPolis project (Grant No. 348480), and by the R3Swarms project funded by the Secure Systems Research Center (SSRC), Technology Innovation Institute (TII).

\bibliographystyle{unsrt}
\bibliography{bibliography}

\end{document}